\documentclass[letterpaper]{article}
\usepackage{aaai16}
\usepackage{times}
\usepackage{helvet}
\usepackage{courier}
\usepackage{setspace}
\usepackage{float}
\usepackage{subcaption}
\usepackage{times}
\usepackage{epsfig}
\usepackage{graphicx}
\usepackage{amsmath}
\usepackage{wrapfig}
\usepackage{amssymb}
\usepackage{caption}
\usepackage{enumitem}
\newcommand{\ignore}[1]{}
\usepackage[noend]{algorithm2e}
\usepackage{algorithmicx,algpseudocode}
\usepackage{booktabs,tabularx}%
\usepackage{multirow} 
\usepackage{amsmath}
\usepackage{amssymb}

\usepackage{epsfig}

\usepackage{url}
\usepackage{latexsym}
\usepackage{graphicx}
\usepackage{amsmath}
\usepackage{amssymb}
\usepackage{subcaption}
\usepackage{epstopdf}
\usepackage{dblfloatfix}
\usepackage{booktabs}
\usepackage{caption}
\usepackage{array}

\newcommand{\quotes}[1]{``#1''}

\usepackage{graphicx}
\usepackage{amsmath,amssymb} 
\usepackage{color}

\def\eg{{e.g. }}

\def\ie{{i.e. }}

\def\ignore#1{}

\def\onedot{.}
\def\eg{\emph{e.g}\onedot~} 
\def\ie{\emph{i.e}\onedot~}

\date{}

\begin{document}

\author{ {\textbf{Mohamed Elhoseiny$^{\S\ddag}$, Jingen Liu$^\ddag$, Hui Cheng$^\ddag$, Harpreet Sawhney$^\ddag$, Ahmed Elgammal$^\S$}} \\ 
\scriptsize{\texttt{m.elhoseiny@cs.rutgers.edu,\{jingen.liu,hui.cheng,harpreet.sawhney\}@sri.com, elgammal@cs.rutgers.edu}} \\
$^\S$Rutgers University, Computer Science Department$\,\, \,\,\,\,\,\,\,\,\,\,\,\,$ $^\ddag$SRI International, Vision and Learning Group 
}
\title{Zero-Shot Event Detection by Multimodal Distributional Semantic Embedding of Videos}


\maketitle

\begin{abstract}

We propose a new zero-shot Event Detection method by Multi-modal Distributional Semantic embedding of videos. Our model embeds object and action concepts as well as other available modalities from videos into a  distributional semantic space. To our knowledge, this is the first Zero-Shot event detection model that is built on top of distributional semantics and extends it in the following directions: (a) semantic embedding of multimodal information in videos (with focus on the visual modalities), (b) automatically determining relevance of concepts/attributes to a free text query, which could be useful for other applications, and (c) retrieving videos by free text event query (e.g., "changing a vehicle tire") based on their content.  We embed videos into  a distributional semantic space and then measure the similarity between videos and the event query in a free text form. We validated our method on the large TRECVID MED  (Multimedia Event Detection) challenge.  Using only the event title as a query, our method outperformed the state-of-the-art that uses big descriptions from 12.6\% to 13.5\% with MAP metric and  0.73 to 0.83 with ROC-AUC metric. It is also an order of magnitude faster. 
\end{abstract}

\section{Introduction}

\begin{figure}[t!]
 \centering
        \begin{subfigure}[b]{0.2\textwidth}
                \includegraphics[width=0.75\textwidth , height=0.8\textwidth]{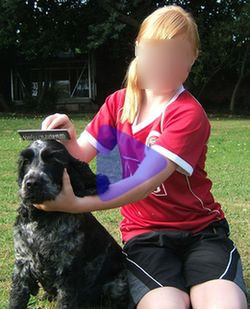}
                \caption{Grooming an Animal  \\ \scriptsize (1) \quotes{brushing dog}, weight = 0.67  \normalsize\\ 
\scriptsize   (2) \quotes{combing dog},  weight = 0.66 \\
\scriptsize (3) \quotes{clipping nails}, weight = 0.52 \normalsize}
                \label{fig:feedanimal}
        \end{subfigure}%
        ~ 
        \begin{subfigure}[b]{0.22\textwidth}
                \includegraphics[width=0.8\textwidth , height=0.5\textwidth]{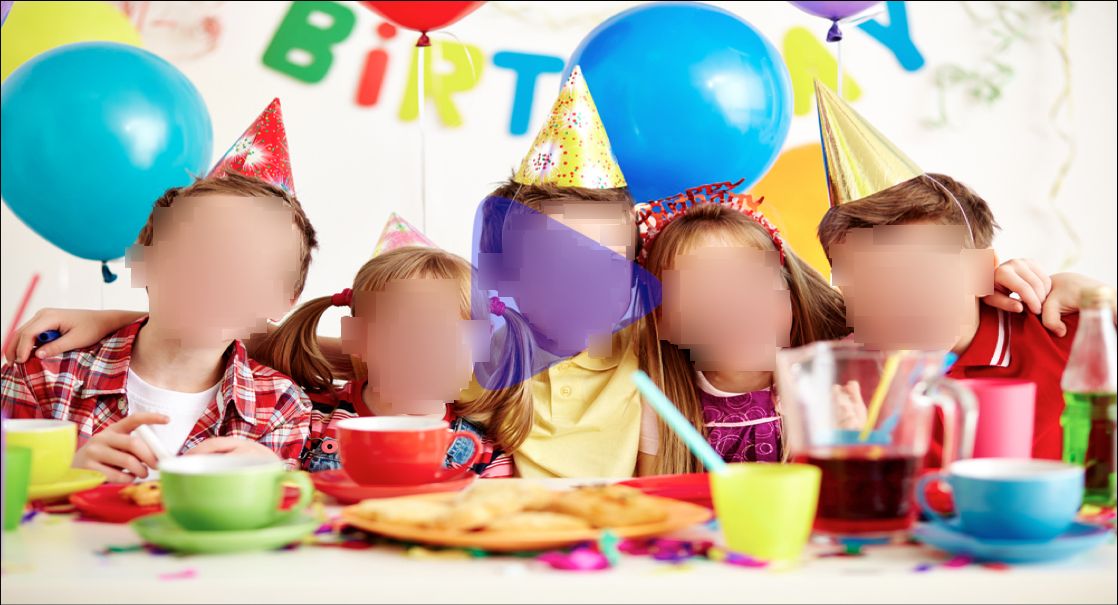}
                \caption{Birthday Party   \\ 
               \scriptsize  (1)  \quotes{cutting cake}, weight = 0.72  \\ \normalsize 
\scriptsize (2) \quotes{blowing candles}, weight = 0.65  \\ \normalsize 
 \scriptsize (3) \quotes{opening presents}, weight = 0.59 \normalsize}
                \label{fig:bdayparty}
        \end{subfigure}
  \caption{Top relevant Concepts from a pre-defined multi-media concept repository and their automatically-assigned  weights  as a part of our Event Detection method}
  \label{figexamev}
\end{figure}

Every minute, hundreds of hours of video are uploaded to video archival site such as YouTube~\cite{ythourscite}. Developing methods to automatically understand the events captured in this  large volume of videos is necessary and meanwhile challenging. One of the important tasks in this direction is event detection in videos. The main objective of  this task is to determine the relevance of a video to an event  based on the video content (e.g., feeding an animal, birthday party; see Fig.~\ref{figexamev}). The cues of an event in a video could include  visual objects, scene, actions, detected speech (by Automated Speech Recognition(ASR)), detected text (by Optical Character Recognition (OCR)), and audio concepts (e.g. music and water concepts).

Search and retrieval of videos for arbitrary events using only free-style text and unseen text in particular has been a dream in computational video and multi-media understanding.  This is referred as \quotes{zero-shot event detection}, because there is no positive exemplar videos to train a detector. Due to the proliferation of videos, especially consumer-generated videos (e.g., YouTube), zero-shot search and retrieval of videos has become an increasingly important problem.

Several research works have been proposed to facilitate performing the zero-shot learning task by establishing an intermediate semantic layer between events or generally categories (i.e., concepts or attributes) and the low-level representation of a multimedia content from the visual perspective.~\cite{lampert2009learning}  and~\cite{farhadi2009describing}  were the two first to use attribute learning representation for the zero-shot setting for object recognition in still images.  Attributes were similarly adopted for recognizing human actions~\cite{liu2011recognizing};  attributes are generalized and denoted by concepts in this context. Later,~\cite{liu2013video} proposed Concept Based Event Retrieval (CBER) for videos InTheWild. Even though these methods facilitate zero-shot  event detection, they only capture the visual modality and more importantly they assume that the relevant concepts for a query event are  manually defined. This manual definition of concepts, also known as semantic query editing, is a tedious task and may be biased due to the limitation of human knowledge.   Instead, we aim at automatically generating relevant concepts by leveraging information  from  distributional semantics.

 Recently, {several systems  were proposed for  zero-shot event detection methods~\cite{Wu2014CVPR,Jiang2014,Jiang_acmM2014,chen2014event,habibian2014composite}.  These approaches rely on the whole text description of an event where relevant concepts are specified; see example event descriptions used in these  approaches in the Supplementary Materials (SM)\footnote{ Supplementary Materials (SM) could be found here \scriptsize{\url{https://sites.google.com/site/mhelhoseiny/EDiSE_supp.zip}}} ( explicitly define the event explication, scene, objects, activities, and  audio). 
In practice, however, we think that typical use of event queries under this setting should be similar to text-search, which is based on few words instead that we model their connection to the multimodal content in  videos.

 { The main question addressed in this paper is how to use an  event text query (i.e. just the event title like \quotes{birthday party} or \quotes{feeding an animal}) to retrieve a ranked list of videos based on their content. In contrast to~\cite{lampert2009learning,liu2013video}, we do not  manually assign relevant concepts for a given event query. Instead, we leverage information from a distributional semantic space~\cite{mikolov2013distributed} trained on a large  text corpus to embed event queries and videos to the same space, where similarity between both could be directly estimated. Furthermore, we only assume that query comes in the form of an \quotes{unstructured} few-keyword query (in contrast to~\cite{Wu2014CVPR,Jiang2014,Jiang_acmM2014}).}   
We abbreviate our method as EDiSE (Event-detection by Distributional Semantic Embedding of videos).

\textbf{Contributions.} The  contributions of this paper can be listed as follows: (1) Studying how to use few-keyword unstructured-text query to detect/retrieve videos based on their multimedia content,  which is novel in this setting.  We show how relevant concepts to that event query could be automatically retrieved through a distributional semantic space and got assigned a weight associated with the relevance; see Fig.~\ref{figexamev} \quotes{Birthday} and \quotes{Grooming an Animal} example events. (2) To the best of our knowledge, our work is the first attempt to model the connection between few keywords and multimodal  information in videos by distributional semantics 
. We study and propose different similarity metrics in the distributional semantic space to enable  event retrieval based on (a) concepts, (b) ASR, and (c) OCR in videos. Our unified framework is capable of embedding all of them into the same space; see Fig.~\ref{figapproach}. 
(3) Our method is also very fast,  which makes it applicable to both large number of   videos and  concepts (\ie 26.67 times  faster than the state of the art~\cite{Jiang_acmM2014}). 


\section{Related Work}
\label{secrelwork}
Attribute methods for zero-shot learning are based on manually specifying the attributes for each category (e.g.,~\cite{lampert2009learning,parikh2011interactively}). Other methods focused on attribute discovery ~\cite{rohrbach2011evaluating,rohrbach2013NIPS} and then apply the same mechanism. 
Recently, several methods were proposed to perform zero shot recognition by representing unstructured text in document terms (\eg ~\cite{elhoseiny2013write,mensink2014costa})
One drawback of the TFIDF~\cite{salton1988term} in~\cite{elhoseiny2013write} and hardly matching tag terms in~\cite{mensink2014costa,rohrbach2010helps} is that they do not capture semantically related terms that our model can relate in noisy videos instead of still images.  Also,  WordNet~\cite{miller1995wordnet}, adopted in ~\cite{rohrbach2010helps}, does not connect objects with actions (e.g., person blowing candle), making it hard to apply in our setting and heavily depending on predefined information like WordNet.

There has been a recent interest especially in the computational linguistics' community in word-vector representation ( e.g., ~\cite{bengio2006neural}), which captures word semantics based on context. While word-vector representation is not new,  recent algorithms (e.g. ~\cite{mikolov2013distributed,mikolov2013efficient}) enabled learning these vectors from billions of words, which makes them much more semantically accurate. As a result, these models got recently adopted in several tasks including translation ~\cite{mikolov2013exploiting} and web search ~\cite{shen2014convolutional}. Several computer vision researchers  explored using these word-vector representation to perform Zero-Shot learning in the object recognition  (e.g.~\cite{frome2013devise,socher2013zero,norouzi2014zero}). They embed the object class name into the word-vector semantic space learnt by models like~\cite{mikolov2013distributed}. It is worth mentioning that these zero-shot learning approaches~\cite{frome2013devise,socher2013zero}  and also the aforementioned work~\cite{elhoseiny2013write} assume that during training, there is a set of training classes and test classes. Hence, they learn a transformation to correlate the information between both domains (textual and visual). In contrast, zero-shot setting of event retrieval rely mainly on the event information without seeing any training events, as assumed in recent zero-shot event retrieval methods (e.g.,  ~\cite{Allan2013,Jiang2014,Wu2014CVPR,liu2013video}). Hence, there does not exist seen events to learn such transformation from. Differently, we also model multimodal connection from free text query to video information.  

In the context of videos,~\cite{Wu2014CVPR} proposed a method for zero-shot event detection by using the salient words in the whole structured event description, where relevant concept  are already defined in the event structured text description; also see Eq. 1 in ~\cite{Wu2014CVPR}. Similarly,~\cite{Allan2013} adopted a Markov-Random-Field language model proposed by~\cite{metzler2005markov}. One drawback of this model is that it performs an intensive processing for each new concept. This is since it determines the relevance of the concept to a query event by creating a text document to represent each  concept. This document is created by web-querying the concept name and some of its keywords and  merging the top retrieved pages. In contrast, our model does not require this step to determine relevance of an event to a query. Once the language model is trained, any concept can be instantly added and captured in our multimodal semantic embedding of videos. 

In contrast to both~\cite{Wu2014CVPR} and~\cite{Allan2013}, we focus on retrieving videos only with the event title (i.e., few-words query) and without semantic editing. The key difference is in modeling and embedding concepts to allow zero-shot event retrieval. In~\cite{Wu2014CVPR} and~\cite{Allan2013}, the semantic space is a vector whose dimensionality is the number of the concepts. Our idea is to embed concepts, video information, and the event query into a distributional semantic space whose dimensionality is “independent” of the number of concepts. This property, together with the semantic properties captured by distributional semantics, feature our approach with two advantages  (a) scalability to any concept size. Having new concepts does not affect the representation dimensionality (i.e., in all our experiments concepts, videos, event queries are embedded to $M$ dimensional space; $M$ is few hundreds in our experiments). (b) facilitating automatic determination of relevant concepts given an unstructured short event query: For example, being able to automatically determine that \quotes{blowing a candle} concept is a relevant concept to \quotes{birthday party} event. ~\cite{Wu2014CVPR} and~\cite{Allan2013}  used the complete text description of an event for retrieval that explicitly specifies relevant concepts.

There is a class of models that improve zero-shot Event Detection performance by reranking. Jiang et al. proposed multimodal pseudo relevance feedback~\cite{Jiang2014} and self-paced reranking~\cite{Jiang_acmM2014} algorithms.  The main assumption behind these models is that all unlabeled test examples are available and the top few examples by a given initial ranking have high top K precision (K $\sim$ 10). This means that reranking algorithms  can not update confidence of a video for an event without knowing the confidences of the remaining videos to perform reranking. In contrast, our goal is different which is to directly model the probability of a few-keyword event-query given an arbitrary video. Hence, our work does not require an initial ranking and can compute the conditional probability of a video  without any information about other videos. Our method is also 26.67  times faster, as detailed   in our experiments.



\section{Method}
\label{secapp}

\subsection{Problem Definition}

Given an arbitrary event query $e$ and a video $v${ (e.g. just "birthday party")}, our objective is to model $p(e|v)$. We start by defining the representation of event query $e$,  the concept set $\mathbf{c}$, the video $v$ in our setting. 

\begin{figure*}
\centering
\hspace{-5mm}
\begin{minipage}{.5\textwidth}
  \centering
  \includegraphics[width=1.0\textwidth]{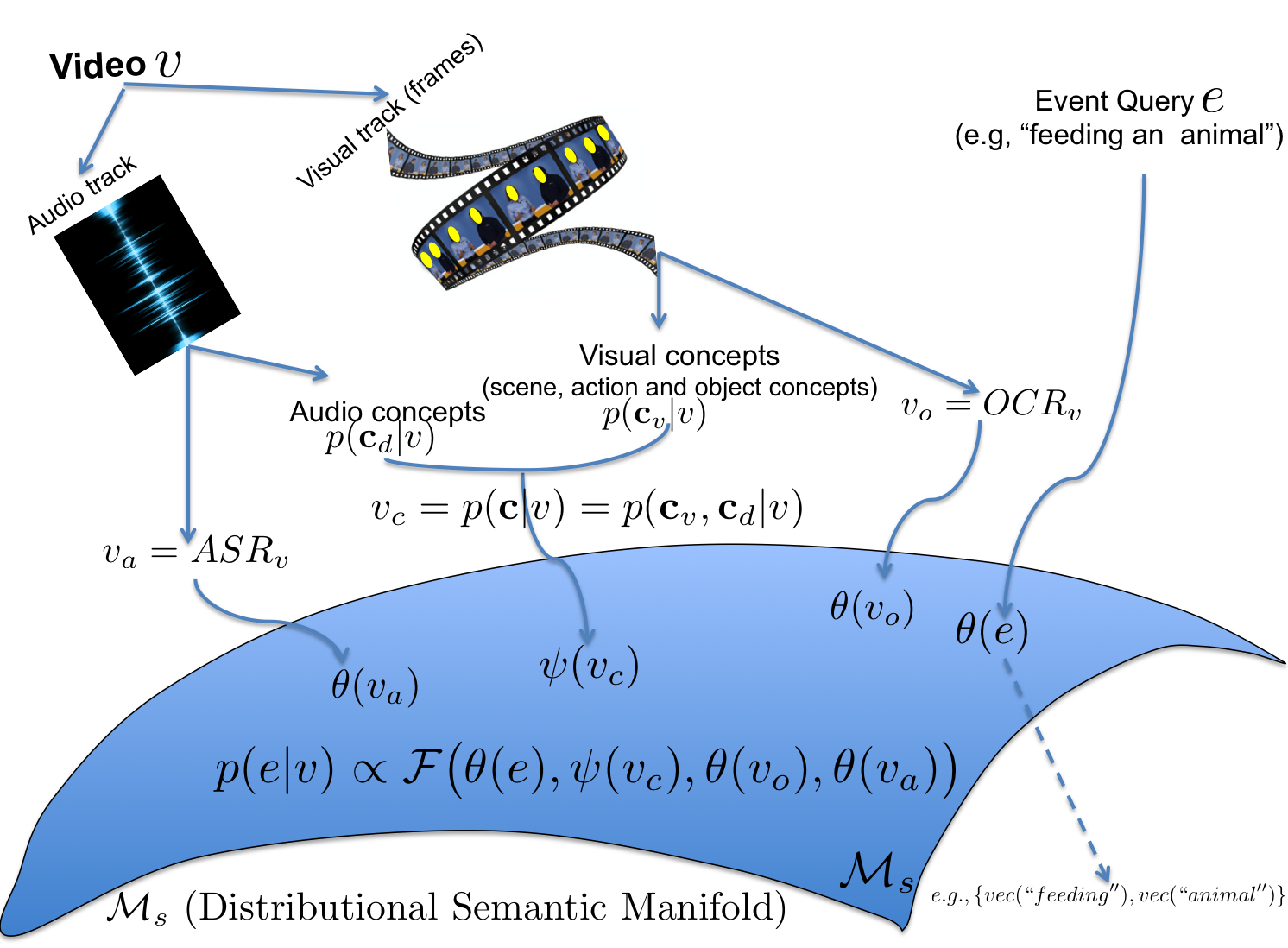}
   \caption{EDiSE Approach}
  \label{figapproach}
\end{minipage}%
\begin{minipage}{.01\textwidth}
$\,$
\end{minipage}
\begin{minipage}{.5\textwidth}
  \centering
  \hspace{-1mm}\includegraphics[width=1.0\textwidth,height=0.60\textwidth]{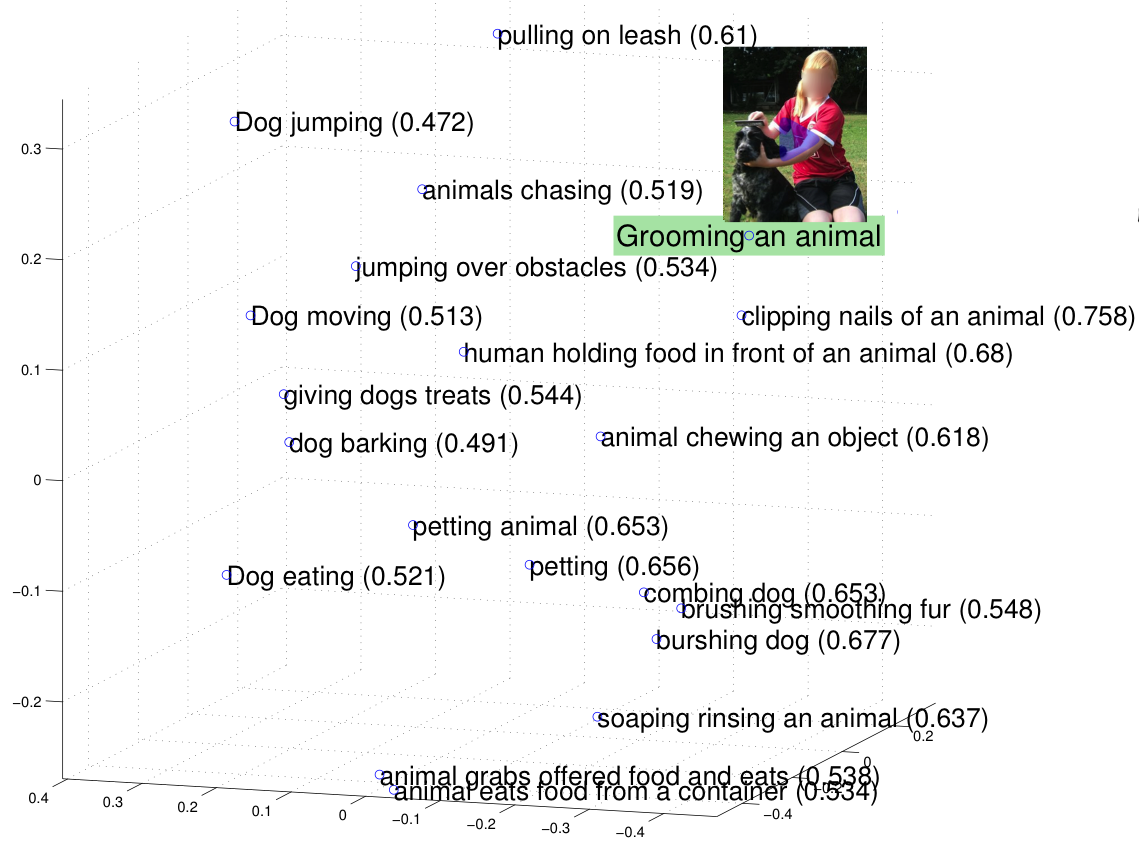}
    \caption{PCA visualization in 3D of the   "Grooming an Animal" event (in green) and its most 20 relevant concepts in $\mathcal{M}_s$ space using $s_p(\cdot,\cdot)$.   The exact $s_p(\theta($\quotes{Grooming An Animal}$), \theta(c_i))$ is shown between parenthesis}
  \label{figdsspaceconcepts}
\end{minipage}
\begin{minipage}{.01\textwidth}
$\,$
\end{minipage}

\end{figure*}

\textbf{Event-Query representation $e$:} We  use the unstructured event title to represent an event query for concept based retrieval. Our framework also allows additional terms specifically for ASR or OCR based retrieval. While we show retrieval on different modalities, concept based retrieval is our main focus in this work. The  few-keyword event query for concept based retrieval is denoted by $e_c$, while query keywords for OCR and ASR are denoted by $e_o$ and $e_a$, respectively. Hence, under our setting $e=\{e_c, e_o, e_a\}$.

\textbf{Concept Set $\textbf{c}$:} We denote the whole concept set in our setting as $\textbf{c}$, which include visual concepts $\textbf{c}_{v}$ and audio concepts $\textbf{c}_{d}$, i.e., $\textbf{c} = \{\textbf{c}_v,\textbf{c}_\textit{{d}}\}$. The visual concepts include object, scene and action concepts.    The audio concepts include acoustic related concepts like water sound. We performed an experiment on a set of audio concepts trained on MFCC audio features~\cite{davis1980comparison,logan2000mel}. However, we found their performance $\approx 1\%$ MAP, and hence we excluded them from our final  experiments. Accordingly, our final performance mainly relies on the visual concepts for concept based retrieval; i.e., $\textbf{c}_d=\emptyset$.  We denote each  member $c_i \in \textbf{c}$ as the definition of the $i^{th}$ concept in $\textbf{c}$. $c_i$ is defined by the $i^{th}$ concept's name and optionally some related keywords; see examples in SM. Hence, $\mathbf{c} = \{c_1, \cdots, c_N\}$  is the the set of concept definitions, where $N$ is the number concepts.

\textbf{Video Representation:} For our zero-shot purpose,  a video $v$ is defined by three pieces of information, which are video OCR denoted by $v_o$, video ASR denoted by $v_a$, and video concept representation denoted by $v_c$.   $v_o$ and $v_a$ are the detected text in OCR and ASR, respectively. We used ~\cite{myers2005rectification} to extract $v_o$ and ~\cite{van2013extracting} to extract $v_a$.  In this paper, we mainly focus on  the visual video content, which is the most challenging. The video concept based representation $v_c$ is defined as 
\begin{equation}
\small
 v_c= [p(c_1|v), p(c_2|v), \cdots , p(c_N|v)]
\end{equation}
where  $p(c_i|v)$ is a conditional probability of  concept $c_i$ given video $v$, detailed later. We denote $p(c_i|v)$ by $v_c^i$.

In zero-shot event detection setting, we aim at recognizing events in videos without training examples based on its multimedia content including   still-image concepts like objects and scenes, action concepts, OCR, and ASR\footnote{Note that OCR and ASR are not concepts. They are rather detected text in video frames and speech}. Given a video $v= \{v_c, v_o, v_a\}$, our  goal is to compute $p(e|v)$ by embedding both the event query $e$ and   information of video $v$ of  different modalities ($v_c$, $v_o$, and $v_o$) into a distributional semantic space, where relevance of $v$ to $e$ could be directly computed; see Fig.~\ref{figapproach}. Specifically,  our approach is to model $p(e|v)$ as a function $\mathcal{F}$ of $\theta(e)$, $\psi(v_c)$,  $\theta({v}_o)$, and $\theta({v}_a)$, which are the distributional semantic embedding of $e$, $v_c$, $v_o$, and $v_a$, respectively
\begin{equation}
\label{eqmain}
\small
\begin{split}
p(e|v) \propto & \mathcal{F}\big( \theta(e),\psi(v_c),\theta(v_o), \theta(v_a) \big) \\
\end{split}
\end{equation}
We remove the stop words from $e$, $v_o$, $v_a$ before applying the embedding $\theta(\cdot)$. The rest of this section is organized as follows. First, we present the distributional semantic manifold and the embedding function $\theta(\cdot)$ which is applied to $e$, $v_a$, $v_o$, and the concept definitions $\mathbf{c}$ in our framework. Then, we show how  to determine automatically relevant concepts to an event title query and  assign a relevance weight to them, as illustrated in Fig.~\ref{figexamev}. We present this concept relevance weighting in a separate section since it might be generally useful  for other applications. Finally, we present the details of $p(e|v)$ where we derive  $v_c$  embedding (\ie $\psi(v_c)$), which is based on the proposed concept relevance weighting.


\subsection{Distributional Semantic Model \& $\theta(\cdot)$ Embedding}
\label{appss1}

We start by the distributional semantic model by~\cite{mikolov2013distributed,mikolov2013efficient} to train our semantic manifold. We denote the trained semantic manifold by $\mathcal{M}_s$, and the vectorization function that maps a word to $\mathcal{M}_s$ space as $vec(\cdot)$. We denote the dimensionality of the real vector returned from $vec(\cdot)$ by $M$. 
 These models  learn a vector for each word $w_n$, such  that $p(w_n|(w_{i-L}, w_{i-L+1}, \cdots,  w_{i+L-1},w_{i+L})$ is maximized over the training corpus;  $2\times L$ is the context window size. Hence similarity between $vec(w_i)$ and $vec(w_j)$ is high if they co-occurred a lot in context of size $2\times L$ in the training text-corpus (i.e., semantically similar words share similar context). Based on the trained $\mathcal{M}_s$ space, we define how to embed  the event query $e$, and $\mathbf{c}$. 
Each of $e_c$, $e_a$, and $e_o$ is set of one or more words.  Each of these words can be directly  embedded into $\mathcal{M}_s$ manifold by $vec(\cdot)$ function. Accordingly, we represent these sets of word vectors for each of $e_c$, $e_a$, and $e_o$  as $\theta(e_c)$, $\theta(e_a)$, and $\theta(e_o)$. We denote $\{\theta(e_c), \theta(e_a), \theta(e_o)\}$ by $\theta(e)$. Regarding embedding of $\mathbf{c}$,  each concept $c^* \in \mathbf{c}$ is defined by its name and optionally some related keywords. Hence,  the corresponding word vectors are then used to define $\theta(c^*)$ in $\mathcal{M}_s$ space. 

\subsection{Relevance of Concepts to Event Query}
\label{appss2}

Let us  define a similarity function between $\theta(c^*)$ and $\theta(e_c)$ as $s(\theta(e_c),\theta(c^*))$. We propose two functions to measure the similarity between $\theta(e_c)$ and  $\theta(c^*)$. The first one is inspired by an example by~\cite{mikolov2013distributed} to show the quality of their language model, where they indicated that $vec($``king''$)+vec($``woman''$)-vec($``man''$)$ is closest to $vec($``queen''$)$. Accordingly, we define a version of $s(X,Y)$, where the sets $X$ and $Y$  are firstly pooled by the sum operation; we denote the sum pooling operation on a set by an overline. For instance, $\overline{X}=\sum_i x_i$ and $\overline{Y} = \sum_i{y_j}$,  where $x_i$ and $y_j$ are the word vectors of the $i^{th}$ element in $X$ and the $j^{th}$ element in $Y$, respectively. Then, cosine similarity between $\overline{X}$ and $\overline{Y}$ is computed. We denote this version as   $s_p(\cdot,\cdot)$; see Eq.~\ref{eqstmerged}. Fig.~\ref{figdsspaceconcepts} shows how $s_p(\cdot,\cdot)$  could be used to retrieve the top 20 concepts relevant to $\theta($\quotes{Grooming An Animal}$)$ in $\mathcal{M}_s$ space. The figure also shows embedding of the query and the relevant concept sets in 3D PCA visualization. $\theta(e_c = $\quotes{Grooming An Animal}$)$ and each of $\theta(c_i)$ for the most relevant 20 concepts are represented by their corresponding pooled vectors ($\overline{\theta(e_c )}$ and  $\overline{\theta(c_i))} \forall i$), normalized to unit length under L2 norm.  Another idea is to define $s(X,Y)$ as a similarity function between the  $X$ and $Y$ sets. For robustness~\cite{torki2010putting}, we used  percentile-based Hausdorff point set metric, where similarity between each pair of points is computed by the cosine similarity. We denote this version by $s_t(X,Y)$; see Eq.~\ref{eqstmerged}. We used $l = 50\%$ (i.e., median).
\begin{equation}
\label{eqstmerged}
  \small
  \begin{split}
    \small
s_p(X,Y)& = \frac{(\sum_i x_i)^\mathsf{T}(\sum_j y_j)}{\|\sum_i x_i\| \|\sum_j y_j\|} = \frac{\overline{X}^\mathsf{T} \overline{Y}}{\|\overline{X}\| \|\overline{Y}\|} \,\,\,\,\,\,\,\,\,\,\,\,\,\,\,\,\,\,\,\,\,\,\,\,\,\,\,\,\,\,\,\,\,\\
  \end{split}
\end{equation}
\begin{equation*}
  \small
  \begin{split}
\,\,s_t(X,Y)& = min\{\underset{j}{\overset{l\%}{min}}\, \underset{i}max \frac{x_i^\mathsf{T} y_j}{\|x_i\| \|\|y_j\|},  \underset{i}{\overset{l\%}{min}}\, \underset{j}max \frac{x_i^\mathsf{T} y_j}{{\|x_i\| \|y_j\|}}\}
  \end{split}
\end{equation*}
%
%
%
%
\subsection{Event Detection $p(e|v)$}
\label{appss3}
In practice, we decomposed $p(e|v)$ into  $p(e_c|v)$, $p(e_o|v)$, $p(e_a|v)$, which makes the problem  reduces to deriving $p(e_c|v)$ (concept based retrieval), $p(e_o|v)$ (OCR based retrieval), and $p(e_a|v)$ (ASR based retrieval) under $\mathcal{M}_s$. We start by $p(e_c|v)$ then we will how later in this section how $p(e_o|v)$, and $p(e_a|v)$ could be estimated.

\textbf{Estimating} ${p}(e_c|v)$ \textbf{:} In our work, concepts are linguistic meanings that have corresponding detection functions given the video $v$. From Fig.~\ref{figdsspaceconcepts}, $\mathcal{M}_s$ space could be viewed as a space of  meanings captured by a training text-corpus,  where only sparse points in that space has a corresponding visual detection functions given  $v$, which are the concepts $\mathbf{c}$ (e.g., \quotes{blowing a candle}). For zero shot event detection, we aim at exploiting these sparse points by the information captured by $s(\theta(e_c),\theta(c^i \in \mathbf{c}))$ in $\mathcal{M}_s$ space. We  derive $p(e_c|v)$ from probabilistic perspective starting from marginalizing $p(e_c|v)$ over the concept set \textbf{c}
\begin{equation}
\label{eqpc}
\small
\begin{split}
p(e_c|v) &\propto \sum_{c_i} p(e_c|{c_i}) p({c}_i|v) \propto  \sum_{c_i} s(\theta(e_c),\theta({c}_i)) v_c^i
\end{split}
\end{equation}
where $p(e|{c}_i) \forall i$ are assumed to be proportional to $s(\theta(e_c),\theta(c_i))$ in our framework.  From semantic embedding perspective, each video $v$ is embedded into $\mathcal{M}_s$ by the set $\psi(v_c) =  \{ \theta_v({c}_i) = v_c^i \theta({c}_i), \forall c_i \in \mathbf{c} \}$, where $v_c^i \theta({c}_i)$ is a set of the same points in $\theta({c}_i)$ scaled with $v_c^i$;  $\psi(v_c)$ could be then directly compared with $\theta(e_c)$; see Eq.~\ref{eqpc2}
\begin{equation}
\label{eqpc2}
\small
\begin{split}
  p(e_c|v)&\propto  \sum_{c_i} s(\theta(e_c),\theta({c}_i)) v_c^i \\
  &\propto s'(\theta(e_c),\psi(v_c)= \{ \theta_{v}({c}_i), \forall c_i \in \mathbf{c} \} ) \\
\end{split}
\end{equation}
where $s'( \theta(e_c), \psi(p(\mathbf{c}|v)) = \sum_i{ s(\theta(e_c), \theta_{v}({c}_i) )}$ and $s(\cdot,\cdot)$ could be replaced by  $s_p(\cdot,\cdot)$, $s_t(\cdot,\cdot)$, or any other measure in $\mathcal{M}_s$ space. An interesting observation is that when $s_p(\cdot, \cdot)$ is chosen, $p(e_c|v) \propto \frac{\overline{\theta(e_c)}^T}{\|\theta{e_c}\| }  \Big( \sum_i \frac{ \overline{\theta(c_i)}}{\|\theta{c_i}\|} v_c^i \Big)$ which is a direct  similarity between $\overline{\theta(e_c)}$ representing the query and the embedding of ${\psi(v_c)}$ as $\sum_i \frac{ \overline{\theta(c_i)}}{\|\theta{c_i}\|} v_c^i$; see proof in Appendix A.      $s_p(\cdot, \cdot)$ performs consistently better than $s_t(\cdot, \cdot)$ in our experiments. In practice, we only include $\theta_{v}({c}_i)$ in  $\psi(v_c)$ such that $c_i$ is among the top R concepts with highest $p(e_c|c_i)$. This is assuming that the remaining concepts are assigned $p(e_c|c_i)=0$ which makes those items vanish; we used R=5. Hence, only a few concept detectors needs to be computed for on $v$ which is a computational advantage. 

\textbf{Estimating}   ${p}(e_o|v)$ and ${p}(e_a|v)$ \textbf{:}
Both ${v}_o$ and ${v}_a$ can be directly embedded into $\mathcal{M}_s$ since they are sets of words. Hence, we can model $p(e_o|v)$ and $p(e_a|v)$ as follows
\begin{equation}
\label{eqpasrocrprpb}
\small
\begin{split}
p(e_o|v) &\propto s_d(\theta(e_o), \theta(v_o) ), 
p(e_a|v) \propto s_d(\theta(e_a)\,\,,\,\, \theta(v_a) ) \\
\end{split}
\end{equation}
where $s_d(X, Y) = \sum_{ij} x_i^T y_j$. We found this similarity function more appropriate for ASR/OCR text since they normally contains more text compared to concept definition.  We also exploited an interesting property in $\mathcal{M}_s$ that nearest words to an arbitrary point can be retrieved. Hence, we automatically augment $e_a$ and $e_o$  with the nearest words to the event title in $\mathcal{M}_s$ using cosine similarity before retrieval. We found this trick effective in practice since it automatically retrieve relevant words that might appear in $v_o$ or $v_a$.

\textbf{Fusion:} We fuse $p(e_c|v)$, $p(e_o|v)$, and $p(e_a|v)$ by weighted geometric mean with focus on visual concepts, i.e.  $p(e|v) =$ \small$ \sqrt[\leftroot{-3}\uproot{3}w+1]{ p(e_c|v)^w \sqrt{p(e_o|v) p(e_a|v))} }$; $w=6$\normalsize.   $p(e_c|v)$, $p(e_c|v)$, and $p(e_c|v)$ involves  the similarity between $\theta(e)$ and each of $\psi(v_c)$, $\theta(v_o)$,  and $\theta(v_a)$,  leading to Eq.~\ref{eqmain} view.

\section{Visual Concept Detection functions ($p(\mathbf{c}|v)$)}
\label{secvisemb}
We leverage the information from three types of visual concepts in $\mathbf{c}_v$:  object concepts  $\mathbf{c}_o$, action concepts $\mathbf{c}_a$, and scene concepts $\mathbf{c}_s$. Hence, $\mathbf{c}=\mathbf{c}_v = \{\mathbf{c}_o \cup \mathbf{c}_a \cup \mathbf{c}_s\}$; the list of concepts are attached in SM. 
We define object and scene concept probabilities per video frame, and action concepts per video chunks. 
The rest of this section summarizes the concept detection for objects and scenes per frame $f$,  and action concepts per video chunk $u$. Then, we show how they can be reduced to video level probabilities.  Fig.~\ref{fig_conceptsscores} shows example high confidence concepts in a \quotes{Birthday Party} video. 

 \begin{figure}[b!]   
  \centering
    \includegraphics[width=0.5\textwidth,height=0.16\textwidth]{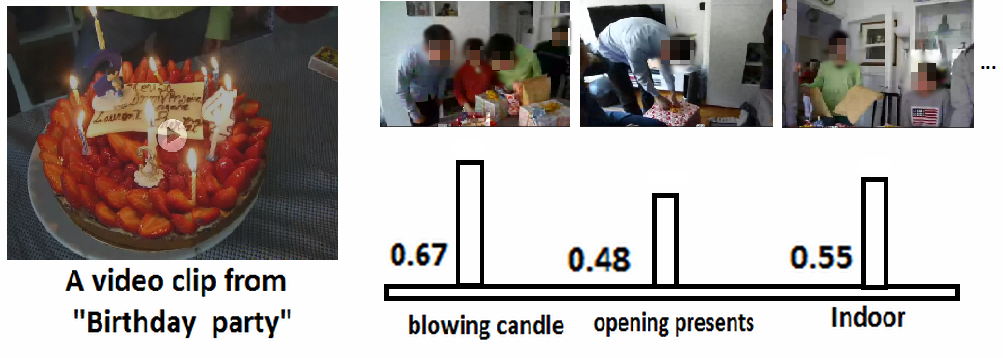}
  \caption{Concept probabilities from videos ($p(\mathbf{c}|v)$)}
  \label{fig_conceptsscores}
\end{figure}}
      
\noindent \textbf{Object Concepts $p(o_i|f), o_i \in \mathbf{c}_o$:} We involve 1000 Overfeat \cite{sermanet2013overfeat} object concept detectors which maps to 1000-ImageNet categories. We also adopt the concept detectors of “face”, “car” and “person”  from a  publicity available detector (i.e., ~\cite{felzenszwalb2008discriminatively})  

\noindent \textbf{Scene Concepts $p(s_i|f), s_i \in \mathbf{c}_s$:} We represented scene concepts ($p(s_i|f)$) as bag of word representation on static features (i.e., SIFT~\cite{lowe2004distinctive} and HOG~\cite{dalal2005histograms}) with 10000 codebooks. We used TRECVID 500 SIN concepts  concepts, including scene categories like \quotes{city} and \quotes{hall} way; these concepts are provided by  provided by TRECVID2011 SIN track.

\noindent  \textbf{Action Concepts $p(a_i|u), a_i \in \mathbf{c}_a$:}
We use both manually annotated (i.e. strongly supervised) and  automatically annotated (i.e. weekly supervised) concepts; detailed in SM. 
 We have $\sim$500 action concepts; please refer to~\cite{liu2013video} for the action concept detection method that we adopt. 

\subsubsection{Video level concept probabilities $p(\mathbf{c}|v)$}
\label{vc_vl_def}
We represent probabilities of the $\mathbf{c}_v$ set given a video $v$ by a pooling operation over the the chunks or the frames of the videos similar to ~\cite{liu2013video}. In our experiments, we evaluated both max and average pooling. Specifically,  $p(o_i|v) = \rho( \{ p(o_i|f_k), f_k \in v \})$, $p(s_l|v) = \rho( \{ p(s_i|f_k), f_k \in v \})$, $p(a_k|v) = \rho( \{ p(a_i|u_k), u_k \in v \}$, 
where $p(o_i|v)$ and $p(s_l|v)$   are the video level probabilities of for the $i^{th}$ object  and the $l^{th}$   scene concepts respectively, pooled over frames $f_k \in v$. $\{f_k \in v\}$ are selected every M frames in $v$ (M= 250).  $p(a_k|v)$ is the video level probability of the $k^{th}$ action concept, pooled over a set of video chunks $\{u_k \in v\}$. The chunk size is set to the mean chunk length of all concept training chunks.  Finally, $\rho$ is the pooling function. We denote average and max pooling as $\rho_{a}(\cdot)$ and $\rho_{m}(\cdot)$,  respectively. 


\section{EDiSE Computational Performance Benefits}
\begin{table*}[hbp!]
\centering
\caption{MED2013 MAP performance on four concept sets (event title query)}
  \scalebox{0.9}
  {
    \begin{tabular}{|c|cc|cc|cc|cc|c|}
    \hline
          & \multicolumn{4}{c|}{Ours-Gnews}     & \multicolumn{4}{c|}{Ours-Wiki}      & (Dalton etal, 2013) \ignore{\cite{Allan2013}} \\
        \cline{2-9}
     TRECVID MED 2013   & \multicolumn{2}{c|}{$\rho_{m}(\cdot)$}   & \multicolumn{2}{c|}{$\rho_{a}(\cdot)$} & \multicolumn{2}{c|}{$\rho_{m}(\cdot)$} & \multicolumn{2}{c|}{$\rho_{a}(\cdot)$} & \textbf{} \\         \cline{2-9}
          & $s_p(\cdot,\cdot)$ & $s_t(\cdot,\cdot)$  & $s_p(\cdot,\cdot)$ & $s_t(\cdot,\cdot)$ & $s_p(\cdot,\cdot)$ & $s_t(\cdot,\cdot)$  & $s_p(\cdot,\cdot)$ & $s_t(\cdot,\cdot)$  &  \\
          \hline
    \textbf{Concepts G1 \ignore{Sarnoff\_AUTO} (152 concepts)} & \textbf{4.29} & 3.94\%  & 2.39\%  & 2.38\%  & 3.14\%  & 2.13\%  & 1.85\%  & 1.70\%  & 2.57\% \\
    \textbf{Concepts G2 \ignore{UCF101} (101 concepts)} & \textbf{1.74} & 1.20  & 1.56\%  & 1.20\%  & 1.09\%  & 0.96\%  & 0.66\%  & 0.60\%  & 1.17\% \\
    \textbf{Concepts G3 \ignore{Sarnoff\_MA} (60 concepts)} & \textbf{1.72} & 1.33\%  & 1.28\%  & 1.16\%  & 1.21\%  & 0.88\%  & 0.88\%  & 0.74\%  & 1.54\% \\
    \textbf{Concepts G4 \ignore{UCF13} (56 concepts)} & \textbf{1.22} & 0.95  & 0.84\%  & 0.69\%  & 0.87\%  & 0.76\%  & 0.67\%  & 0.56\%  & 0.83\% \\
\hline
    \end{tabular}%

    }
  \label{tab:tblumasscomp}
\end{table*}

Here we discuss the computational complexity of concept based EDISE, and ASR/OCR based EDiSE. The fusion part is negligible since it is constant time.

\subsection{Concept based EDiSE}
The computational complexity for computing $p(e_c|v)$ is mainly linear in the number of videos, denoted by $|V|$. We here detail why computational complexity of $p(e_c|v)$ is almost constant and hence video retrieval is almost $O(|V|)$. 

From Eq.~\ref{eqpc2}, $p(e_c|v)$  has a computational complexity of $O(N \cdot Q)$ for on e video, where $Q$ is the computational complexity of computing $s(\cdot, \cdot)$ and $N$ is the number of concepts. We detail next the computational complexity of $s_p(\cdot, \cdot)$ and $s_t(\cdot, \cdot)$  for the whole set of videos $|V|$.

\subsubsection{ Complexity of $p(e_c|v)$ for $s_p(\cdot, \cdot)$}

Let's assume that there $\theta(e_c)$ set has $|e_c|$ terms  and $\theta(c_i)$ has $|c_i|$ terms. Then, the computational complexity of $s_p(\theta(e_c), \theta(c_i))$ is $O(M (|e_c| + |c_i|)$. $|c_i|$ and  and $|e_c|$ are usually few terms in our case ($<10$). Hence the computational complexity of $s_p(\theta(e_c), \theta(c_i))$ is $O(M)$, where $M$ is the dimensionality of the word vectors. In our experiments $M=300$. Given the complexity of  $s_p(\theta(e_c), \theta(c_i))$, the computational complexity of $p(e_c|v)$ will be $O(N \cdot M)$, where $N$ is the number of concepts.  Hence, the computational complexity for computing  $p(e_c|v)$ for $|V|$ videos is $O(|V| \cdot N \cdot M)$. However, for a given event, only few concepts are relevant, which are computed based on $s_p(\theta(e_c), \theta(c_i))$ and only few concepts $5$ in our case are sufficient for event zero shot retrieval, retrieved by Nearest  Neighbor search of $c_i\in \mathbf{c}$ that is close the $e_c$. Hence the computational complexity reduced to $O(|V| \cdot M)$, $M=300$ for the GoogleNews word2vec model that we used. Hence, the complexity for $|V|$ videos is basically linear $O(|V|)$, given $M$ is a constant and  $M <<|V|$.

\subsubsection{ Complexity of $p(e_c|v)$ for $s_t(\cdot, \cdot)$}

The previous argument applies here in all elements except the complexity of the similarity function  $s_t(\theta(e_c), \theta(c_i))$, which is $O(M (|e_c| \cdot |c_i|)$. Assuming that $|e_c| \cdot |c_i|$ is bounded by a constant, then the complexity of $|V|$ videos is also $O(|V| \cdot M)$, but with a bigger constant compared to $s_p(\cdot, \cdot)$ (linear in $|V|$ for constant $M<<|V|$).

\subsection{ASR/OCR based EDiSE}

The computational complexity of $s_d(\theta(e_o), \theta(v_o))$  and $s_d(\theta(e_a), \theta(v_a))$  are  $O(|e_o| \cdot |v_o| \cdot M)$ and $O(|e_a| \cdot |v_a| \cdot M)$, respectively. There is no concepts for ASR/OCR based retrieval. Hence, the computational complexity of $p(e_o, v)$ and $p(e_a|v)$ are $O(|V|\cdot |e_o| \cdot |v_o| \cdot M)$ and $O(|V|\cdot |e_a| \cdot |v_a| \cdot M)$, respectively. Since $|e_o| \ll |V|$, $|v_o| \ll |V|$, $|e_a| \ll |V|$, $|v_a| \ll |V|$, and  $M \ll |V|$, the  dominating factor in the complexity for both $p(e_o, v)$ and $p(e_a|v)$ will be $|V|$.

\section{Experiments}
\label{secexpr}

We evaluated our method on the large TRECVID MED ~\cite{trecvid13}. We show the MAP (Mean Average Precision) and ROC AUC performance of  the designated MEDTest set~\cite{trecvid13},  containing more than 25,000   videos\ignore{\footnote{Information about the data splits and the events are detailed i~\cite{trecvid13}}}. Unless otherwise mentioned, our results are on TRECVID MED2013. There are two distributional semantic models in our experiments, trained on Wikipedia and GoogleNews using~\cite{mikolov2013distributed}. The Wikipedia model got trained on 1 billion words resulting in a vocabulary of size of$\approx$120,000 words and word vectors of 250 dimensions. The GoogleNews model got trained on 100 billion words resulting in a vocabulary of size 3 million words and word vectors of 300 dimensions. The objective of having two models is to compare how well our EDiSE method  performs depending on the size of the training corpus, used to train the language model.  In the rest of this section, we present Concepts, OCR, ASR, and fusion results. 

\subsection{Concept based Retrieval}

All the results in this section were generated by automatically retrieved concepts using only the event title. We start by comparing different settings of our method against~\cite{Allan2013}. We used the language model in~\cite{Allan2013} for concept based retrieval to rank the concepts. This indicates that $p(e|c_i)$ in Eq.~\ref{eqpc} is computed by the language model in ~\cite{metzler2005markov} as adopted in ~\cite{Allan2013}, that we compare with under exactly the same setting. For our model, we evaluated the two pooling operations $\rho_{m}(\cdot)$ and $\rho_{p}(\cdot)$ and also the two different similarity measures on $\mathcal{M}_s$ space $s_p(\cdot,\cdot)$ and $s_t(\cdot,\cdot)$. Furthermore, we evaluated the methods on both Wikipedia and GNews language models. In order to have conclusive experiments on these eight settings of our model compared to ~\cite{Allan2013}, we performed all of them   on the  four  different sets of concepts (i.e. each has the same concept detectors; completely consistent comparison); see Table~\ref{tab:tblumasscomp}. Details about these concept sets are attached in SM.

\begin{figure*}[b!]
        \centering
        \begin{subfigure}[b]{0.61\textwidth}
                 \includegraphics[width=1.0\textwidth, height=0.32\textwidth]{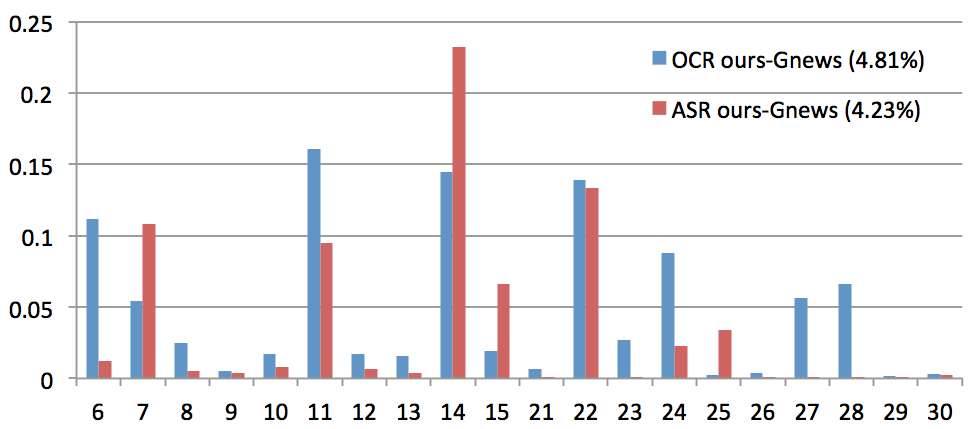}
                \caption{MED2013}
                \label{fig:med13asrocr}
        \end{subfigure}%
        ~ 
        \begin{subfigure}[b]{0.36\textwidth}
               \includegraphics[width=1.0\textwidth, height=0.53\textwidth]{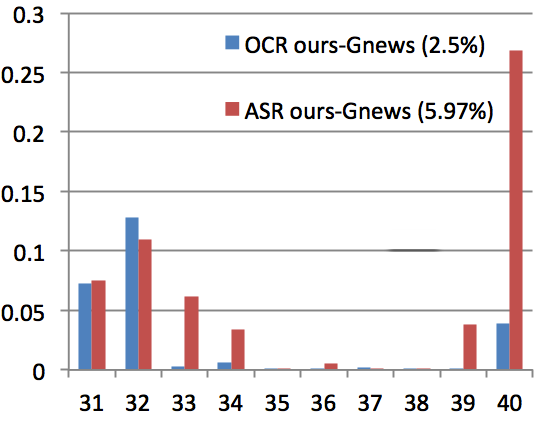}
                \caption{MED2014 (E31:40)}
                \label{fig:med141to40asrocr}
        \end{subfigure}
          \caption{ASR \& OCR AP Performance (Google News)}
        \label{fig:asroceMAP}
\end{figure*}
\begin{figure*}[b!]
  \centering
  \includegraphics[width=0.80\textwidth, height=0.235\textwidth]{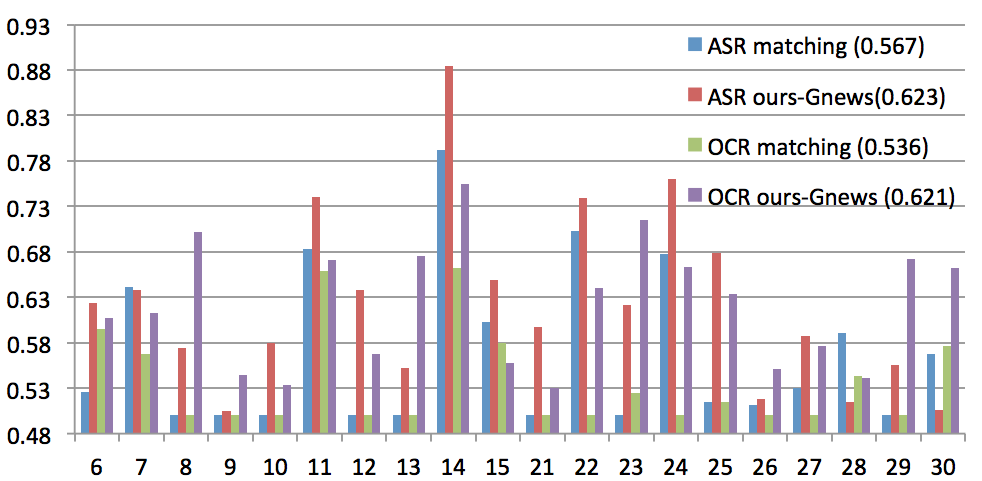}
  \caption{ASR \& OCR AUCs on MED2013: Ours (GoogleNews) vs keyword Matching (the same  query)}
  \label{figaucocrasr}
\end{figure*}

There are a number of observations. (1) using GNews (the bigger text corpus) language model is consistently better than  using the Wikipedia language model. This indicates  when the word embedding model is trained with a bigger text corpus, it captures more semantics and hence more accurate in our setting. (2) max pooling $\rho_{m}(\cdot)$ behaves consistently better than average pooling $\rho_{a}(\cdot)$. (3) $s_p(\cdot,\cdot)$ similarity measure is  consistently better than $s_t(\cdot,\cdot)$, which we see very interesting since this indicates that our hypothesis of using the vector operations on $\mathcal{M}_s$ manifold better represent $p(e|c_i)$. Hence, we recommend finally to use the model trained on the larger corpus,  $\rho_{m}(\cdot)$ for concept pooling, and use $s_p(\cdot,\cdot)$ to measure the performance on $\mathcal{M}_s$ manifold. (4) our model's final setting is consistently better than~\cite{Allan2013}. The final MED13 ROC AUC performance is 0.834. MAP for MED13 Events 31 to 40 (E31:40) is 5.97\%. Detailed figures are attached in SM.

\begin{table}[t!]
  \caption{MED2013 full concept set MAP Performance (auto-weighted versus manually-weighted concepts)}
    \scalebox{0.8}
    {
    \begin{tabular}{|c|c|c|c|}
    \hline
    {Ours (auto)}&{Dalton etal,13(auto)}& {Dalton etal,13(manual)}& Overfeat\ignore{~\cite{sermanet2013overfeat}} \\
    \hline
    \textbf{8.36\%} & {3.40\%} & 7.4\%&  2.43\%\\
    \hline
    \end{tabular}%
    }
    \scalebox{0.8}
    {
    \begin{tabular}{|c|c|c|c|c|}
\hline
    {SUN\ignore{~\cite{patterson2012sun,Wu2014CVPR}}} & {Object Rank\ignore{~\cite{li2010object,Wu2014CVPR}}} & {Classeme\ignore{~\cite{torresani2010efficient,Wu2014CVPR}}} & ${CD}^{DT}$\ignore{~\cite{Wu2014CVPR}}&  \small $WSC^{D-SIFT}_{YouTube}$ \normalsize \ignore{ ~\cite{Wu2014CVPR} }\\
\hline
    {0.48\%} & {0.77\%} &  0.84\% & 2.28\% & 3.48\%\\
\hline
    \end{tabular}%
    }
  \label{tab:soacomp}%
\end{table}%
Our next experiment shows the final MAP performance using the recommended setting for our framework on the whole set of concepts, detailed  earlier and in SM. Table~\ref{tab:soacomp} shows our final performance compared with~\cite{Allan2013} on the same concept detectors. It is not hard to see that our method performs more than double the MAP performance of~\cite{Allan2013} under the same concept set. Even when manual semantic editing is applied to~\cite{Allan2013}, our performance is  still better without semantic editing. We also show the performance on the same events of different concepts (i.e. SUN~\cite{patterson2012sun}, Object Rank~\cite{li2010object}, Classeme~\cite{torresani2010efficient}), and the best performing concepts in ~\cite{Wu2014CVPR} (i.e., ${CD}^{DT}$, $WSC^{D-SIFT}_{YouTube}$). {These numbers are as reported in~\cite{Wu2014CVPR}}. The results indicate the value of our  concepts  and approach compared to~\cite{Wu2014CVPR} and their concepts. We also report our performance using Overfeat concepts only to retrieve videos for the same events.  This shows the value of involving action and scene concepts {compared} to only still image concepts like Overfeat for zero-shot event detection. We highlight that the results  in~\cite{Wu2014CVPR} uses the whole event description which explicitly includes names of relevant concepts.

\subsection{ASR and OCR based Retrieval}


First, we compared our OCR  and ASR retrieval trained on both Wikipedia and GoogleNews language model. Table~\ref{tab:asrocrwikiGoog} shows that the GoogleNews model MED13 MAP is better than the Wikipedia Model  MAP in both ASR and OCR, which is  consistent with our concept retrieval results.  Fig.~\ref{fig:asroceMAP} shows the GoogleNews MED13 AP per event for both OCR and ASR. We further show our AP performance on MED14 events 31 to 40 in Fig.~\ref{fig:asroceMAP}.

\begin{table}[bp!]
  \centering
  \caption{ASR \& OCR Retrieval MAP on $\mathcal{M}_s$ using GNews, Wikipedia, and using  word matching}
  \scalebox{0.8}{
    \begin{tabular}{|c|c|c|c|}
  \hline
    & {GNews MED2013}  & {Wiki MED2013}  & {matching MED2013} \\
    \hline
   OCR &\textbf{4.81\%} & {3.85\%}  & {1.8\%} \\ \hline
  ASR &     \textbf{4.23\%} & {1.50\%} & {3.77\%}  \\
    \hline
    \end{tabular}%
    }
  \label{tab:asrocrwikiGoog}%
\end{table}%
\begin{table}[bp!] 
  \caption{ASR \& OCR MAP performance using GNews corpus compared to~\cite{Wu2014CVPR}(prefix E indicates Event)}
  \label{tblocr}
  \scalebox{0.8}{
    \begin{tabular}{|c|c|c|c|}
    \hline
           & {MED13 } & {MED13~\cite{Wu2014CVPR}}  & {MED14(E31:40)} \\
    \hline
    {OCR}  & \textbf{4.81\%} & {4.30\%}  & {2.5\%} \\     \hline
    \end{tabular}%
    }
  \scalebox{0.8}{
    \begin{tabular}{|c|c|c|c|}
    \hline
          &  MED13 & MED13~\cite{Wu2014CVPR}  & { MED14 (E31:40)}\\
    \hline
    {ASR} & \textbf{4.23\%} & {3.66\%} &  {5.97\%}  \\
    \hline
    \end{tabular}%
  
    }
  \label{tabasrocr}%
\end{table}%

In order to show the value our semantic modeling, we  computed the performance of string matching method  as a baseline, which basically increment the score for every exact match in the the detected text to words in the query. While, both our model and the matching model use the same query words and ASR/OCR detection, semantic properties captured by$~\mathcal{M}_s$ boosts the performance compared to string matching; see table~\ref{tab:asrocrwikiGoog}. This is since semantically relevant terms to the query have a high cosine similarity in $\mathcal{M}_s$ (i.e., high $vec(w_i)^\mathsf{T} vec(w_j)$ if $w_i$ is semantically related to $w_j$). On the other hand,  hard matching basically assumes that $vec(w_i)^\mathsf{T} vec(w_j)=1$ if$ w_i=w_j$, 0 otherwise. We also computed the ROC AUC metric for our method and the hard matching method on ASR and OCR; see Fig.~\ref{figaucocrasr}.  For ASR, average AUC is 0.623 for ours and 0.567 for Matching (9.9\% gain). For OCR, average AUC is 0.621 for ours and 0.53 for Matching (17.1\% gain). We  report our GNews model results compared with~\cite{Wu2014CVPR} to indicate that, we achieve state-of-the-art MED13 MAP performance or even better for ASR/OCR; see table~\ref{tabasrocr}. The table also shows our ASR\&OCR MED14 (E31:40) MAP.

\subsection{Fusion Experiments and Related Systems}

In table~\ref{tblfusion}, we start by presenting a summary of our earlier ASR/OCR results on MED13 Test. Comparing OCR and ASR performances to Concepts performance, it is not hard to see that OCR/ASR have much lower average AUC zero-shot performance compared to concepts which are visual in our work. This indicates that OCR/ASR produces much higher false negatives compared to visual concepts.  When we fused our all OCR and ASR confidences, we achieved 10.7\% MAP performance, however, the average AUC performance is as low as 0.67. We achieved lower MAP for our concepts 8.36\% MAP but the average AUC performance is as high as 0.834. This indicates that measuring retrieval performance on MAP performance only is not informative, so one approach might achieve a high MAP but lower average AUC and vice versa. We further achieved the best performance of our system by fusing all Concepts, OCR, and ASR  to achieve 13.1\% MAP and 0.830 average AUC. We found our system achieves better than the state of the art system~\cite{Wu2014CVPR} 4.0\% gain in MAP, but significantly in average AUC; see 13.6\% gain to~\cite{Wu2014CVPR} in table~\ref{tblfusion}.

We also discuss  CPRF~\cite{yang2010supervised}, MMPRF~\cite{Jiang2014},  and SPaR~\cite{Jiang_acmM2014} reranking systems in contrast to our system that does not involve reranking. The initial retrieval performance is  3.9\% MAP without reranking. Interestingly, we achieved a performance of 13.1\% MAP also without reranking. The reranking methods assumes high top 5-10 precision of the initial ranking and that all test videos are available.\ignore{ Then, the top examples are likely positives and negative examples are selected at random based on the assumption.} Without any of these assumptions,  our system without reranking performs 6.7\%,  3.0\%, and 0.2\% better than CPRF~\cite{yang2010supervised}, MMPRF~\cite{Jiang2014}, and SPaR~\cite{Jiang_acmM2014} re-ranking systems; see table~\ref{tblfusion}. Unfortunately, ROC AUC performances are not available for these method to compare with. Regarding efficiency, given   $v_c$ representation of videos, our concept retrieval experiment on our whole concept set it takes $\approx$270 seconds on a 16 cores Intel Xeon processor (64GB RAM) to the retrieval task on 20 events altogether. This is more than the time that SPaR~\cite{Jiang_acmM2014}  takes to rerank  one event on an Intel Xeon processor(16GB RAM); see ~\cite{Jiang_acmM2014}. Since, we detect the MED13 events in $\approx$270 given $v_c$ representation of videos and as reported in~\cite{Jiang_acmM2014}, their average detection time per event for MED13 is $\approx$ 5 minutes assuming feature representation of videos (i.e., 360 seconds per event = 7200 seconds per 20 events). This indicates that our system is 26.67X faster than~\cite{Jiang_acmM2014} in MED13 detection. Finally, when we applied SPaR on our output as an initial ranking, we found that it improves MAP (from 13.1\% to 13.5\%) but hurts ROC AUC (from 0.83 to 0.79). This indicates that reranking has a limited/harmful effect on the performance of our method. We think is since our method already achieve a high  performance without re-ranking; see SM for details about the features in this experiment.   

\begin{table} [t!]
\caption{Fusion Experiments and Comparison to State of the Art Systems}
\label{tblfusion}
\centering
\scalebox{0.8}
{
\begin{tabular}{|c|c|c|}
\hline 
Method & MAP & AUC \\ 
\hline 
Our Concept retrieval (event title query) & 8.36\% & 0.834  \\
 Concept retrieval (Dalton etal, 2013) (event title query) & 3.4 \% & - \\ 
  Concept retrieval (Dalton etal, 2013) (manual concepts) & 7.4\% & -
\\  \hline 
Our ASR  GNews&  4.81\%&  0.623\\ 
Our OCR GNews& 4.23\% & 0.621 \\ 
Our ASR  Matching& 2.77\% & 0.567 \\ 
Our OCR Matching& 1.8\% & 0.536 \\ \hline 
Our ASR and OCR all fused& 10.6 & 0.670 \\ 
\hline 
\textbf{Our Full (Concepts+ASR+OCR) (No reranking)}  & \textbf{13.1\% }& \textbf{0.830} \\ 
\textbf{Our Full  + SPaR reranking~\cite{Jiang_acmM2014}}  & \textbf{13.5\% }& \textbf{0.790} \\ 
\hline 
Full system \cite{Wu2014CVPR}  & 12.6 &  0.730 \\
\hline
\multicolumn{3}{|c|}{\textbf{Reranking Systems}} \\ \hline 
Without  Reranking~\cite{Jiang2014}& 3.9\% & -  \\ 
CPRF~\cite{yang2010supervised}& 6.4\% & -  \\
Full Reranking system MMPRF\cite{Jiang2014}& 10.1\% & -  \\ 
Full Reranking system SPaR\cite{Jiang_acmM2014}  & 12.9\% & - 
 \\  
\hline 
\end{tabular} 
}
\end{table}

\section{Conclusion}
\label{secconc}

We proposed a method for zero shot event detection by distributional semantic embedding of video modalities and with only event title  query. By fusing all modalities, our method outperformed the state of the art on the challenging TRECVID MED benchmark. Based on this notion, we also showed how to automatically determine relevance of concepts to an event based on the distributional semantic space. 

\vspace{2mm}

\noindent\textbf{Acknowledgements.} This work has been supported by the Intelligence Advanced Research Projects Activity (IARPA) via Department of Interior National Business Center contract number D11-PC20066. The U.S. Government is authorized to reproduce and distribute reprints for Governmental purposes
notwithstanding any copyright annotation thereon. The views and conclusions contained herein are those of the authors and should not be interpreted as necessarily representing the official policies or endorsements, either expressed or implied, of IARPA, DOI/NBC, or the U.S. Government. This work is also partially funded by  NSF-USA award \# 1409683.

{\small
\bibliographystyle{aaai}
\bibliography{egbib2.bib}

\begin{thebibliography}{}

\bibitem[\protect\citeauthoryear{Bengio \bgroup et al\mbox.\egroup
  }{2006}]{bengio2006neural}
Bengio, Y.; Schwenk, H.; Sen{\'e}cal, J.-S.; Morin, F.; and Gauvain, J.-L.
\newblock 2006.
\newblock Neural probabilistic language models.
\newblock In {\em Innovations in Machine Learning}.
\newblock  137--186.

\bibitem[\protect\citeauthoryear{Chen \bgroup et al\mbox.\egroup
  }{2014}]{chen2014event}
Chen, J.; Cui, Y.; Ye, G.; Liu, D.; and Chang, S.-F.
\newblock 2014.
\newblock Event-driven semantic concept discovery by exploiting weakly tagged
  internet images.
\newblock In {\em ICMR}.

\bibitem[\protect\citeauthoryear{Dalal and Triggs}{2005}]{dalal2005histograms}
Dalal, N., and Triggs, B.
\newblock 2005.
\newblock Histograms of oriented gradients for human detection.
\newblock In {\em CVPR}.

\bibitem[\protect\citeauthoryear{Dalton, Allan, and Mirajkar}{2013}]{Allan2013}
Dalton, J.; Allan, J.; and Mirajkar, P.
\newblock 2013.
\newblock Zero-shot video retrieval using content and concepts.
\newblock In {\em CIKM}.

\bibitem[\protect\citeauthoryear{Davis and
  Mermelstein}{1980}]{davis1980comparison}
Davis, S., and Mermelstein, P.
\newblock 1980.
\newblock Comparison of parametric representations for monosyllabic word
  recognition in continuously spoken sentences.
\newblock {\em Acoustics, Speech and Signal Processing, IEEE Transactions on}
  28(4):357--366.

\bibitem[\protect\citeauthoryear{Elhoseiny, Saleh, and
  Elgammal}{2013}]{elhoseiny2013write}
Elhoseiny, M.; Saleh, B.; and Elgammal, A.
\newblock 2013.
\newblock Write a classifier: Zero-shot learning using purely textual
  descriptions.
\newblock In {\em ICCV}.

\bibitem[\protect\citeauthoryear{Farhadi \bgroup et al\mbox.\egroup
  }{2009}]{farhadi2009describing}
Farhadi, A.; Endres, I.; Hoiem, D.; and Forsyth, D.
\newblock 2009.
\newblock Describing objects by their attributes.
\newblock In {\em CVPR}.

\bibitem[\protect\citeauthoryear{Felzenszwalb, McAllester, and
  Ramanan}{2008}]{felzenszwalb2008discriminatively}
Felzenszwalb, P.; McAllester, D.; and Ramanan, D.
\newblock 2008.
\newblock A discriminatively trained, multiscale, deformable part model.
\newblock In {\em CVPR}.

\bibitem[\protect\citeauthoryear{Felzenszwalb, McAllester, and
  Ramanan}{2013}]{trecvid13}
Felzenszwalb, P.; McAllester, D.; and Ramanan, D.
\newblock 2013.
\newblock Trecvid 2013 – an overview of the goals, tasks, data, evaluation
  mechanisms and metrics.
\newblock In {\em TRECVID 2013},  1--8.
\newblock NIST.

\bibitem[\protect\citeauthoryear{Frome \bgroup et al\mbox.\egroup
  }{2013}]{frome2013devise}
Frome, A.; Corrado, G.~S.; Shlens, J.; Bengio, S.; Dean, J.; Mikolov, T.;
  et~al.
\newblock 2013.
\newblock Devise: A deep visual-semantic embedding model.
\newblock In {\em NIPS}.

\bibitem[\protect\citeauthoryear{Google2014}{}]{ythourscite}
Google2014.
\newblock {Youtube}, howpublished =
  "\url{https://www.youtube.com/yt/press/statistics.html}", year = {2014}, note
  = "[online; 11/06/2014]".

\bibitem[\protect\citeauthoryear{Habibian, Mensink, and
  Snoek}{2014}]{habibian2014composite}
Habibian, A.; Mensink, T.; and Snoek, C.~G.
\newblock 2014.
\newblock Composite concept discovery for zero-shot video event detection.
\newblock In {\em ICMR}.

\bibitem[\protect\citeauthoryear{Jiang \bgroup et al\mbox.\egroup
  }{2014a}]{Jiang_acmM2014}
Jiang, L.; Meng, D.; Mitamura, T.; and Hauptmann, A.~G.
\newblock 2014a.
\newblock Easy samples first: Self-paced reranking for zero-example multimedia
  search.
\newblock In {\em ACM Multimedia}.

\bibitem[\protect\citeauthoryear{Jiang \bgroup et al\mbox.\egroup
  }{2014b}]{Jiang2014}
Jiang, L.; Mitamura, T.; Yu, S.-I.; and Hauptmann, A.~G.
\newblock 2014b.
\newblock Zero-example event search using multimodal pseudo relevance feedback.
\newblock In {\em ICMR}.

\bibitem[\protect\citeauthoryear{Lampert, Nickisch, and
  Harmeling}{2009}]{lampert2009learning}
Lampert, C.~H.; Nickisch, H.; and Harmeling, S.
\newblock 2009.
\newblock Learning to detect unseen object classes by between-class attribute
  transfer.
\newblock In {\em CVPR}.

\bibitem[\protect\citeauthoryear{Li \bgroup et al\mbox.\egroup
  }{2010}]{li2010object}
Li, L.-J.; Su, H.; Fei-Fei, L.; and Xing, E.~P.
\newblock 2010.
\newblock Object bank: A high-level image representation for scene
  classification \& semantic feature sparsification.
\newblock In {\em NIPS}.

\bibitem[\protect\citeauthoryear{Liu \bgroup et al\mbox.\egroup
  }{2013}]{liu2013video}
Liu, J.; Yu, Q.; Javed, O.; Ali, S.; Tamrakar, A.; Divakaran, A.; Cheng, H.;
  and Sawhney, H.
\newblock 2013.
\newblock Video event recognition using concept attributes.
\newblock In {\em WACV}.

\bibitem[\protect\citeauthoryear{Liu, Kuipers, and
  Savarese}{2011}]{liu2011recognizing}
Liu, J.; Kuipers, B.; and Savarese, S.
\newblock 2011.
\newblock Recognizing human actions by attributes.
\newblock In {\em CVPR}.

\bibitem[\protect\citeauthoryear{Logan and others}{2000}]{logan2000mel}
Logan, B., et~al.
\newblock 2000.
\newblock Mel frequency cepstral coefficients for music modeling.
\newblock In {\em ISMIR}.

\bibitem[\protect\citeauthoryear{Lowe}{2004}]{lowe2004distinctive}
Lowe, D.~G.
\newblock 2004.
\newblock Distinctive image features from scale-invariant keypoints.
\newblock {\em IJCV}.

\bibitem[\protect\citeauthoryear{Mensink, Gavves, and
  Snoek}{2014}]{mensink2014costa}
Mensink, T.; Gavves, E.; and Snoek, C.~G.
\newblock 2014.
\newblock Costa: Co-occurrence statistics for zero-shot classification.
\newblock In {\em CVPR}.

\bibitem[\protect\citeauthoryear{Metzler and Croft}{2005}]{metzler2005markov}
Metzler, D., and Croft, W.~B.
\newblock 2005.
\newblock A markov random field model for term dependencies.
\newblock In {\em ACM SIGIR}.

\bibitem[\protect\citeauthoryear{Mikolov \bgroup et al\mbox.\egroup
  }{2013a}]{mikolov2013efficient}
Mikolov, T.; Chen, K.; Corrado, G.; and Dean, J.
\newblock 2013a.
\newblock Efficient estimation of word representations in vector space.
\newblock {\em ICLR}.

\bibitem[\protect\citeauthoryear{Mikolov \bgroup et al\mbox.\egroup
  }{2013b}]{mikolov2013distributed}
Mikolov, T.; Sutskever, I.; Chen, K.; Corrado, G.~S.; and Dean, J.
\newblock 2013b.
\newblock Distributed representations of words and phrases and their
  compositionality.
\newblock In {\em NIPS}.

\bibitem[\protect\citeauthoryear{Mikolov, Le, and
  Sutskever}{2013}]{mikolov2013exploiting}
Mikolov, T.; Le, Q.~V.; and Sutskever, I.
\newblock 2013.
\newblock Exploiting similarities among languages for machine translation.
\newblock {\em arXiv preprint arXiv:1309.4168}.

\bibitem[\protect\citeauthoryear{Miller}{1995}]{miller1995wordnet}
Miller, G.~A.
\newblock 1995.
\newblock Wordnet: a lexical database for english.
\newblock {\em Communications of the ACM} 38(11):39--41.

\bibitem[\protect\citeauthoryear{Myers \bgroup et al\mbox.\egroup
  }{2005}]{myers2005rectification}
Myers, G.~K.; Bolles, R.~C.; Luong, Q.-T.; Herson, J.~A.; and Aradhye, H.~B.
\newblock 2005.
\newblock Rectification and recognition of text in 3-d scenes.
\newblock {\em IJDAR} 7.

\bibitem[\protect\citeauthoryear{Norouzi \bgroup et al\mbox.\egroup
  }{2014}]{norouzi2014zero}
Norouzi, M.; Mikolov, T.; Bengio, S.; Singer, Y.; Shlens, J.; Frome, A.;
  Corrado, G.~S.; and Dean, J.
\newblock 2014.
\newblock Zero-shot learning by convex combination of semantic embeddings.
\newblock In {\em ICLR}.

\bibitem[\protect\citeauthoryear{Parikh and
  Grauman}{2011}]{parikh2011interactively}
Parikh, D., and Grauman, K.
\newblock 2011.
\newblock Interactively building a discriminative vocabulary of nameable
  attributes.
\newblock In {\em CVPR}.

\bibitem[\protect\citeauthoryear{Patterson and Hays}{2012}]{patterson2012sun}
Patterson, G., and Hays, J.
\newblock 2012.
\newblock Sun attribute database: Discovering, annotating, and recognizing
  scene attributes.
\newblock In {\em CVPR}.

\bibitem[\protect\citeauthoryear{Rohrbach \bgroup et al\mbox.\egroup
  }{2010}]{rohrbach2010helps}
Rohrbach, M.; Stark, M.; Szarvas, G.; Gurevych, I.; and Schiele, B.
\newblock 2010.
\newblock What helps where--and why? semantic relatedness for knowledge
  transfer.
\newblock In {\em CVPR}.

\bibitem[\protect\citeauthoryear{Rohrbach, Ebert, and
  Schiele}{2013}]{rohrbach2013NIPS}
Rohrbach, M.; Ebert, S.; and Schiele, B.
\newblock 2013.
\newblock Transfer learning in a transductive setting.
\newblock In {\em NIPS}.

\bibitem[\protect\citeauthoryear{Rohrbach, Stark, and
  Schiele}{2011}]{rohrbach2011evaluating}
Rohrbach, M.; Stark, M.; and Schiele, B.
\newblock 2011.
\newblock Evaluating knowledge transfer and zero-shot learning in a large-scale
  setting.
\newblock In {\em CVPR}.

\bibitem[\protect\citeauthoryear{Salton and Buckley}{1988}]{salton1988term}
Salton, G., and Buckley, C.
\newblock 1988.
\newblock Term-weighting approaches in automatic text retrieval.
\newblock {\em Information processing \& management}.

\bibitem[\protect\citeauthoryear{Sermanet \bgroup et al\mbox.\egroup
  }{2014}]{sermanet2013overfeat}
Sermanet, P.; Eigen, D.; Zhang, X.; Mathieu, M.; Fergus, R.; and LeCun, Y.
\newblock 2014.
\newblock Overfeat: Integrated recognition, localization and detection using
  convolutional networks.
\newblock In {\em ICLR}.

\bibitem[\protect\citeauthoryear{Shen \bgroup et al\mbox.\egroup
  }{2014}]{shen2014convolutional}
Shen, Y.; He, X.; Gao, J.; Deng, L.; and Mesnil, G.
\newblock 2014.
\newblock A convolutional latent semantic model for web search.
\newblock Technical report, Technical Report MSR-TR-2014-55, Microsoft
  Research.

\bibitem[\protect\citeauthoryear{Socher \bgroup et al\mbox.\egroup
  }{2013}]{socher2013zero}
Socher, R.; Ganjoo, M.; Manning, C.~D.; and Ng, A.
\newblock 2013.
\newblock Zero-shot learning through cross-modal transfer.
\newblock In {\em NIPS}.

\bibitem[\protect\citeauthoryear{Torki and Elgammal}{2010}]{torki2010putting}
Torki, M., and Elgammal, A.
\newblock 2010.
\newblock Putting local features on a manifold.
\newblock In {\em CVPR}.

\bibitem[\protect\citeauthoryear{Torresani, Szummer, and
  Fitzgibbon}{2010}]{torresani2010efficient}
Torresani, L.; Szummer, M.; and Fitzgibbon, A.
\newblock 2010.
\newblock Efficient object category recognition using classemes.
\newblock In {\em ECCV}.

\bibitem[\protect\citeauthoryear{van Hout \bgroup et al\mbox.\egroup
  }{2013}]{van2013extracting}
van Hout, J.; Akbacak, M.; Castan, D.; Yeh, E.; and Sanchez, M.
\newblock 2013.
\newblock Extracting spoken and acoustic concepts for multimedia event
  detection.
\newblock In {\em ICASSP}.

\bibitem[\protect\citeauthoryear{Wu \bgroup et al\mbox.\egroup
  }{2014}]{Wu2014CVPR}
Wu, S.; Bondugula, S.; Luisier, F.; Zhuang, X.; and Natarajan, P.
\newblock 2014.
\newblock Zero-shot event detection using multi-modal fusion of weakly
  supervised concepts.
\newblock In {\em CVPR}.

\bibitem[\protect\citeauthoryear{Yang and Hanjalic}{2010}]{yang2010supervised}
Yang, L., and Hanjalic, A.
\newblock 2010.
\newblock Supervised reranking for web image search.
\newblock In {\em ACM Multimedia}.

\end{thebibliography}
}

\section*{Appendix A: Proof $p(e_c|v)$ for $s(\cdot,\cdot) = s_p(\cdot,\cdot)$}

We start by Eq.~\ref{eqpc2} while replacing $s(\cdot,\cdot)$ as $s_p(\cdot,\cdot)$.
\begin{equation}
\begin{split}
p &(e_c|v)\propto \sum_i s_p(\theta(e_c), \theta(c_i)) p(c_i|v) \\
& \propto \sum_i \frac{\overline{\theta(e_c)}^T \overline{\theta(c_i)}}{\|\theta{e_c}\| \|\theta{c_i}\|} v_c^i  \propto \frac{\overline{\theta(e_c)}^T}{\|\theta{e_c}\| }  \Big( \sum_i \frac{ \overline{\theta(c_i)}}{\|\theta{c_i}\|} v_c^i \Big)
\end{split}
\end{equation}
which is the dot product between $ \frac{\overline{\theta(e_c)}^T}{\|\theta{e_c}\| }$ representing the event embedding, and  $\sum_i \frac{ \overline{\theta(c_i)}}{\|\theta{c_i}\|} v_c^i$ representing the video embedding, which is a function of $\psi(v_c^i) = \{ \theta_v(c_i) =   \theta(c_i) v_c^i \}$. This equation clarifies our  notion of distributional semantic embedding of videos and relating it to event title
\end{document}